\documentclass[10pt,twocolumn,letterpaper]{article}

\usepackage[pagenumbers]{cvpr} % To force page numbers, e.g. for an arXiv version

\usepackage[dvipsnames]{xcolor}

\usepackage{bbding}
\usepackage{pifont}
\usepackage{bm}

\usepackage{multirow}

\definecolor{cvprblue}{rgb}{0.21,0.49,0.74}
\usepackage[pagebackref,breaklinks,colorlinks,citecolor=cvprblue]{hyperref}

\def\paperID{3285} % *** Enter the Paper ID here
\def\confName{CVPR}
\def\confYear{2024}

\title{DiffusionTalker: Personalization and Acceleration for Speech-Driven 3D Face Diffuser}

\author{Peng Chen$^{1,2*}$, Xiaobao Wei$^{1,2*}$, Ming Lu$^{3}$, Yitong Zhu$^{1,2}$, \\ Naiming Yao$^{1}$, Xingyu Xiao$^{4}$, Hui Chen$^{1\dagger}$ \\
$^1$Institute of Software, Chinese Academy of Sciences \\
$^2$University of Chinese Academy of Sciences \\
$^3$Intel Labs China \ \ $^4$Tsinghua University \\
\texttt{chenpeng23@mails.ucas.ac.cn}
}

\begin{document}
\maketitle
\begin{abstract}
% The ABSTRACT is to be in fully justified italicized text, at the top of the left-hand column, below the author and affiliation information.
% Use the word ``Abstract'' as the title, in 12-point Times, boldface type, centered relative to the column, initially capitalized.
% The abstract is to be in 10-point, single-spaced type.
% Leave two blank lines after the Abstract, then begin the main text.
% Look at previous \confName abstracts to get a feel for style and length.

Speech-driven 3D facial animation has been an attractive task in both academia and industry. Traditional methods mostly focus on learning a deterministic mapping from speech to animation. Recent approaches start to consider the non-deterministic fact of speech-driven 3D face animation and employ the diffusion model for the task. However, personalizing facial animation and accelerating animation generation are still two major limitations of existing diffusion-based methods. To address the above limitations, we propose DiffusionTalker, a diffusion-based method that utilizes contrastive learning to personalize 3D facial animation and knowledge distillation to accelerate 3D animation generation. Specifically, to enable personalization, we introduce a learnable talking identity to aggregate knowledge in audio sequences. The proposed identity embeddings extract customized facial cues across different people in a contrastive learning manner. During inference, users can obtain personalized facial animation based on input audio, reflecting a specific talking style. With a trained diffusion model with hundreds of steps, we distill it into a lightweight model with 8 steps for acceleration. Extensive experiments are conducted to demonstrate that our method outperforms state-of-the-art methods. The code will be released. We have provided the supplementary video: \href{https://chenvoid.github.io/DiffusionTalker/}{https://chenvoid.github.io/DiffusionTalker/}
\end{abstract}

\renewcommand{\thefootnote}{\fnsymbol{footnote}} 
\footnotetext[1]{Equal Contribution.} 
\footnotetext[2]{Corresponding Author.}

\section{Introduction}

\begin{figure}
    \centering
    \includegraphics[width=0.8\linewidth]{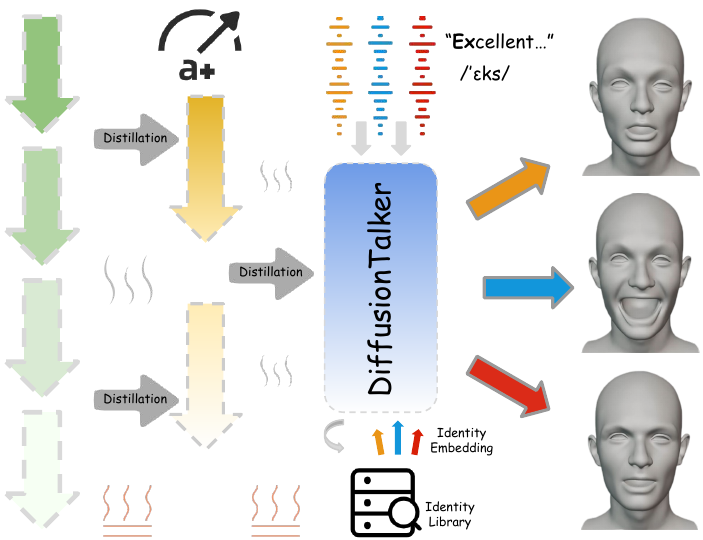}
    \caption{\textbf{DiffusionTalker.} We reduce the steps of the diffusion model for faster inference by knowledge distillation. Based on the model with fewer steps, given an audio sequence, we can find a matching identity embedding in the identity library to personalize the speaker's talking style.}
    \label{fig:intro.pdf}
\end{figure}

Speech-driven 3D facial animation is a crucial task in virtual reality~\cite{wohlgenannt2020virtual}, augmented reality~\cite{chollet2009multimodal} and computer gaming~\cite{ping2013computer, edwards2016jali, zhou2018visemenet} applications. 
% However, 3D facial animation synthesis is time-consuming and requires experienced animators. 
To achieve high-fidelity facial animation, animators need to iteratively adjust facial parameters to fit target performers. With the development of deep learning, end-to-end speech-driven facial animation synthesis has been widely explored. Recent works focus on deep learning-based 3D facial animation, which can produce more vivid animation and facilitate cost reduction. Based on the convolution neural network~\cite{cudeiro2019capture, richard2021meshtalk} or transformer~\cite{vaswani2017attention, fan2022faceformer,peng2023emotalk}, these works try to learn a mapping between audio input and mesh or blendshape~\cite{lewis2014practice} output. Traditional methods mainly focus on learning a deterministic mapping from speech to animation~\cite{cudeiro2019capture, richard2021meshtalk, fan2022faceformer,peng2023emotalk}. Recently, the diffusion model has shown promising potential in learning non-deterministic mapping~\cite{ho2020denoising, song2020denoising, rombach2022high, stan2023facediffuser}. In other words, when strictly controlling lip movements~\cite{sheng2022deep}, the upper face, which has weak correlation with speech, will exhibit a greater diversity of animations. It is the inherent non-deterministic facial cues dispersed across the face.
% Given the inherent non-deterministic facial cues dispersed across the face, speech-driven 3D facial animation aligns seamlessly with the capabilities of the diffusion model. 

However, recent methods still have two major limitations. First, they suffer from a prolonged inference time~\cite{richard2021meshtalk, peng2023emotalk, stan2023facediffuser}, which constitutes a critical limitation in terms of engineering applications. Secondly, in real life, people can recognize a familiar person simply by listening to his voice, discerning his identity based on unique vocal characteristics without seeing his face. Therefore, different individuals exhibit distinct speaking characteristics~\cite{argente1992speech, prajwal2020learning}, indicating the importance of personalization for speech-driven 3D facial animation. 

To address these issues, we propose a novel diffusion-based approach named DiffusionTalker, for producing high-quality 3D facial animations based on the user's speaking characteristics in a short time, as shown in Fig.~\ref{fig:intro.pdf}. For the personalization capacity, we introduce novel identity embeddings corresponding to the speaking characteristics of the training people. We then train the identity encoder to encode embeddings to identity features, which respond to audio features in a contrastive learning manner~\cite{radford2021learning, tian2020contrastive}. 
% We design a contrast loss between learnable identity features and audio features. 
For the synthesis acceleration, we utilize knowledge distillation~\cite{salimans2022progressive} to distill knowledge from teacher DiffusionTalker with enormous steps to student DiffusionTalker with few steps. To leverage dense information from dense sampling steps, we obtain student DiffusionTalker with a progressive training scheme. During inference, with the voice from a new performer, DiffusionTalker first extracts features from the voice and searches for the best corresponding identity embedding. Then, all features and the noise are concatenated and sent to the GRU-based~\cite{cho2014learning} talker decoder for denoising. In such an end-to-end learning pipeline, DiffusionTalker can generate high-fidelity 3D facial animation efficiently. Our main contributions are summarized as follows:

\begin{itemize}
\item We introduce DiffusionTalker, an end-to-end neural network for generating personalized speech-driven 3D facial animation based on Denoising Diffusion Probabilistic Model (DDPM)~\cite{ho2020denoising}. We first apply contrastive learning to get personalized animation, by pulling encoded audio features and learnable identity embeddings together while pushing the mismatched farther apart.

\item To reduce the time cost and maintain high-fidelity generation, we utilize knowledge distillation for DiffusionTalker. After the distillation process, DiffusionTalker generates animation efficiently and obtains 65.5 times acceleration with only 8 time steps.

\item Experimental results show that our DiffusionTalker outperforms the state-of-the-art methods with 0.0969 average lip error while maintaining adequate diversities. 
\end{itemize}
\section{Related Work}

\textbf{Speech-driven 3D Facial Animation.} 
Methods for speech-driven 3D facial animation can be divided into two branches: phoneme-based methods~\cite{charalambous2019audio} and data-driven based methods~\cite{pham2018end, taylor2017deep, huang2018visual}. Phoneme-based methods, such as JALI~\cite{edwards2016jali, zhou2018visemenet}, come with the advantage of easy integration into other artists' pipelines. However, they require intermediary representations of phonemes for co-articulation. Different from phoneme-based methods, data-driven based methods can learn the mapping between audio and facial animation automatically. VOCA ~\cite{cudeiro2019capture} is the first work that uses CNN to map voice to 3D mesh. They also propose a one-hot-style embedding for a controllable output. MeshTalk~\cite{richard2021meshtalk} tries to disentanglement the audio and 3D animation in a categorical latent space, but consumes longer time with less than 100ms of audio as input. FaceFormer~\cite{fan2022faceformer} is the first work to utilize an autoregressive transformer as the model backbone and applies the attention mechanism to extract audio context features. EmoTalk~\cite{peng2023emotalk} focuses on integrating emotion into FaceFormer for better facial animation with a cross-attention module~\cite{vaswani2017attention}. Upon methods, all learn a deterministic mapping between audio and facial animation, while neglecting the non-deterministic nature. FaceDiffuser~\cite{stan2023facediffuser} first attempts to apply the diffusion model to the task, but lacks the ability to personalized output. To achieve personalized facial animation, DiffusionTalker utilizes contrastive learning-based identity embeddings to store personal information.

\noindent
\textbf{Personalization with Diffusion Model.} Personalization is a crucial technique in the research field of diffusion models. One notable approach is DreamBooth \cite{ruiz2023dreambooth}. This method presents a novel pathway for the personalization of text-to-image diffusion models. Given a few images of a particular subject, the pre-trained text-to-image model can be fine-tuned to associate a unique identifier with that subject. Kumari et al. \cite{kumari2023multi} also find that optimizing a few parameters in diffusion models can represent new concepts of the given images. They further propose to combine multiple concepts via closed-form constrained optimization. Liu et al.\cite{liu2023cones} find that a small portion of the neurons can correspond to a particular subject. They use the statistics of network gradients to identify the neurons. Concatenating multiple clusters of concept neurons can vividly generate all related concepts in a single image. Han et al.\cite{han2023svdiff} propose to fine-tune the singular values of the weight matrices, leading to a compact and efficient parameter space for personalization. 
% A few recent works study the protection against malicious use of personalization techniques. For example, \cite{van2023anti} adds subtle noise perturbations to the images for personalization before publishing. 
In our work, we integrated the contrastive learning~\cite{radford2021learning} for matching of audio and identity modalities into the diffusion model training to achieve personalization of speaking styles based on input audio.

\begin{figure*}[h]
\vspace{-10pt}
    \centering
    \includegraphics[width=1.0\textwidth]{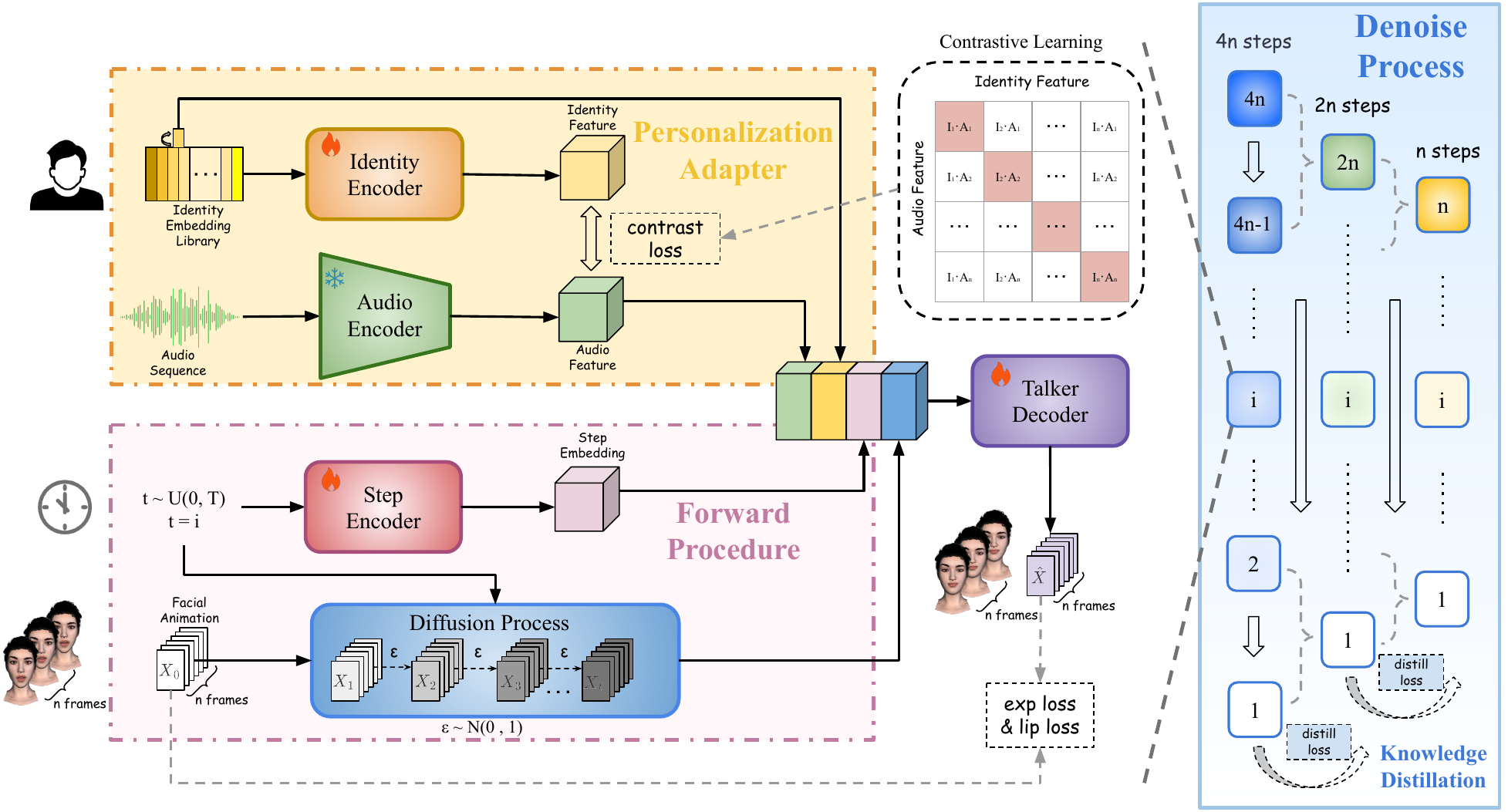}
    \vspace{-5pt}
    \caption{\textbf{Pipeline of DiffusionTalker.}
    DiffusionTalker continuously removes Gaussian noise from noise-added facial animation during the Denoise Process while updating the model's parameters to generate facial animation based on input speech. In each step, the model consists of two parts: the personalization adapter, which uses contrastive learning to match speech and identity features, and the forward procedure, which adds noise to the facial parameters. Finally, all the information is fed into the talker decoder to predict facial animation. DiffusionTalker employs a training approach which is knowledge distillation and reduces the number of sampling steps by half.}
    \label{fig:method}
    \vspace{-10pt}
\end{figure*}

\noindent
\textbf{Distillation for Diffusion Acceleration.} Diffusion models have demonstrated excellent potential for content generation~\cite{rombach2022high, stan2023facediffuser, chen2023diffusiondet, ji2023ddp}, and outperforming GANs~\cite{goodfellow2014generative} and autoregressive models~\cite{brown2020language, oord2016wavenet}. However, they suffer from slow generation due to iterative denoising. Previous studies have suggested the application of knowledge distillation to improve inference speed~\cite{hinton2015distilling}. Specifically, a faster student model can be trained to replicate the output of a pretrained teacher diffusion model. Meng et al.\cite{meng2023distillation} first train a student model to mimic the outputs of a pre-trained model, followed by a progressive distillation. Li et al.\cite{li2023snapfusion} introduce an optimized UNet architecture~\cite{ronneberger2015u, oktay2018attention} that identifies and eliminates redundancies within the original model. It employs data distillation to streamline computational tasks in the image decoder and improves the efficiency of step distillation by incorporating advanced training strategies and regularization informed by classifier-free guidance. BOOT~\cite{gu2023boot} circumvents the dependency on extensive offline computation using a data-free distillation algorithm, significantly enhancing the efficiency of synthetic data generation. Salimans et al.~\cite{salimans2022progressive} propose a progressive distillation approach, halving the number of required sampling steps each time.
% TRACT~\cite{berthelot2023tract} advances the methodology of binary time-distillation to minimize the number of network calls for a given architecture, contributing to the overall reduction of computational demands during inference.
We adopt this approach in our work, using a pre-trained large-step DiffusionTalker as a teacher model to distill a small-step student model. The aim is to accelerate the inference while also enhancing the generalization ability of the student model through knowledge distillation.

% \begin{figure*}[h]
% \vspace{-10pt}
%     \centering
%     \includegraphics[width=1.0\textwidth]{figs/pipeline_v4.pdf}
%     \vspace{-5pt}
%     \caption{\textbf{Pipeline of DiffusionTalker.}
%     DiffusionTalker continuously removes Gaussian noise from facial animation during the Denoise Process while updating the model's internal weights to generate facial animation based on input speech. In each step, the model consists of two parts: the Personalization Adapter, which uses contrastive learning to match speech and identity features, and the Forward Procedure, which adds noise to the model. Finally, all of this information is fed into the Talker Decoder to predict facial animation. DiffusionTalker employs a training approach that incorporates knowledge distillation and reduces the number of steps by half.}
%     \label{fig:method}
%     \vspace{-10pt}
% \end{figure*}

\section{Methodology}

In this section, we first briefly review the diffusion model. Subsequently, we proceed to elaborate on the pipeline of the proposed DiffusionTalker. Finally, we further elucidate the novel designs introduced in DiffusionTalker.

\subsection{Preliminary}
Denoising Diffusion Probabilistic Models (DDPMs)~\cite{ho2020denoising} have become a pivotal element in content generation~\cite{chen2020wavegrad, nichol2021glide}. The fundamental function of DDPMs is to learn the distribution of training data and generate images that closely match this distribution. The DDPM process consists of two stages: the diffusion process and the denoise process. In the diffusion process, noise is added incrementally to the image, gradually transforming it into pure noise. The diffusion process can be characterized by a series of variance variables \( \alpha_1, \alpha_2, \ldots, \alpha_T \), which scale the data distribution at each step \( t \) through the diffusion sequence. The latent variable at any step \( t \) can be expressed as a function of the initial data \( \bm{x}_0 \) and noise \( \bm{\epsilon} \), such that:
\begin{equation}
  \bm{x}_t = \sqrt{\bar{\alpha_t}} \bm{x}_0 + \sqrt{1 - \bar{\alpha_t}} \bm{\epsilon}
\end{equation}
where \( \bm{x}_0 \) denotes the initial data distribution and \( \bm{x}_t \) represents the denoised image at step \( t \). \( \bm{\epsilon} \) is noise sampled from a standard Gaussian distribution \( \mathcal{N}(0, I) \). \( q(\bm{x}_t|\bm{x}_{t-1}) \) is the conditional distribution of \( \bm{x}_t \) given \( \bm{x}_{t-1} \), which describes the process of adding Gaussian noise to \( \bm{x}_{t-1} \), resulting in a noisier latent variable \( \bm{x}_t \) at each step. The variables \( \beta_1, \beta_2, \ldots, \beta_T \) correspond to the noise levels added at each step of the diffusion process. These variables are related to the variance variables through the relationships as follows.
\begin{equation}
  \alpha_t = 1 - \beta_t
\end{equation}
\begin{equation}
  \bar{\alpha_t} = \alpha_1 \alpha_2 \ldots \alpha_t
\end{equation}

The denoise process is designed to reverse the diffusion process by predicting \(\bm{x}_{t-1}\) from \(\bm{x}_t\). The goal is to approximate the true data distribution \(p(\bm{x}_{t-1}|\bm{x}_t)\) starting from the noise distribution. By optimizing the parameters \(\theta\), the denoise process uses a parameterized model \(p_\theta(\bm{x}_{t-1}|\bm{x}_t)\) to closely approximate the true conditional distribution \(p(\bm{x}_{t-1}|\bm{x}_t)\), and ultimately, the data distribution \(p(\bm{x})\).

During DDPM training, the objective at each step is to accurately predict the noise sampled from a Gaussian distribution. Consequently, the training loss can be formulated as follows:
\begin{equation}
    loss = ||\bm{\epsilon} - \epsilon_\theta(\sqrt{\bar{\alpha_t}}\bm{x}_0 + \sqrt{1 - \bar{\alpha_t}}\bm{\epsilon}, t)||^2
\end{equation}
where $\bm{\epsilon}$ is Gaussian noise, $\epsilon_\theta$ is the model that predicts t-step Gaussian noise and $\theta$ is the parameters of it.

During inference, $\bm{x}_{t-1}$ can be represented as follows:
\begin{equation}
\bm{x}_{t-1} = \frac{1}{\sqrt{\alpha_t}} \left( \bm{x}_t - \frac{1 - \alpha_t}{\sqrt{1 - \alpha_t}} \epsilon_\theta(\bm{x}_t, t) \right) + \sigma_t \mathbf{z}
\end{equation}
where $\sigma_t \mathbf{z}$ is a random noise term to generate diversity.
  
%\begin{itemize}
%  \item A random latent variable \( x_t \) is sampled from the distribution \( q(x_0) \).
%  \item A random timestep is chosen, represented by \( t \sim \text{Uniform}(1, \ldots, T) \).
%  \item A noise vector is sampled from a standard Gaussian distribution, denoted by \( \epsilon \sim \mathcal{N}(0, I) \).
%  \item The loss is computed as:
%  \begin{equation}
%    \text{loss} = \epsilon - \epsilon_\theta(\sqrt{\bar{\alpha_t}}x_0 + \sqrt{1 - \bar{\alpha_t}}\epsilon, t)
%  \end{equation}
%  \item Parameter optimization is performed through backpropagation to minimize this loss, which improves the model's ability to denoise and thus better reconstruct the data.
%\end{itemize}

\subsection{Overall Pipeline}
DiffusionTalker is trained based on the DDPM, utilizing audio and identity as conditions to guide the denoise process of facial animation synthesis, thereby generating speech-driven 3D facial animation, as shown in Fig.~\ref{fig:method}. The specific formula employed in this process is as follows:

\begin{equation}
    \bm{\hat{x}}
 = DiffusionTalker_{\theta}(\bm{a}, \bm{i}, \bm{x}_{t}, t)
\end{equation}
where $\theta$ indicates model parameters in DiffsionTalker, $\bm{a}$ is the input audio sequence, $\bm{i}$ is the identity embedding from the identity embedding library, $\bm{x}_{t}$ is $\bm{x}_{0}$ after t steps of the DDPM noise addition process and $\bm{\hat{x}}$ is the predicted facial animation which will be fitted to the ground truth $\bm{x}_{0}$. 
% $\bm{X_{t}}$, obtained after t steps of the DDPM noise addition process, is defined as follows:

% \begin{equation}
%     \bm{X_{t}}
%  = \sqrt{\overline{\alpha}_t \bm{X_0}} + \sqrt{1 - \overline{\alpha}_t \bm{\epsilon}} 
% \end{equation}
% where $\overline{\alpha}_t$ is the hyperparameter of the reparameterization, and $\bm{\epsilon}$ is Gaussian noise following a Gaussian distribution.

In the denoise process of DiffusionTalker, Gaussian noise is continuously removed from noise-added facial animation, and model parameters are updated to predict facial animation based on speech. Each step in this process includes two key components: the personalization adapter and the forward procedure. The former includes an identity embedding library, where each embedding corresponds to an audio sequence. The identity embedding and the audio sequence are encoded using an identity encoder and an audio encoder, respectively. Contrastive learning~\cite{wu2018unsupervised, chen2020simple, tian2020contrastive} is then applied to match the features of both, allowing unknown input audio to find a similar identity embedding in the identity library, thus achieving personalization of the talking style during inference. The identity embedding library can be continuously enriched and expanded by fine-tuning DiffusionTalker. The forward procedure involves the noise addition process, where Gaussian noise is added to the ground truth for $t$ steps, with each step encoded by a step encoder. Finally, all components are fed into the talker decoder to predict facial animation parameters. To accelerate inference speed, we trained DiffusionTalker using knowledge distillation. This approach allows distilling a teacher model with $2n$ steps into a student model with $n$ steps, accelerating the speed of speech-driven 3D face animation synthesis. We will introduce the modules in the following sections.

\subsection{Personalization Adapter}

\begin{figure}[t]
  \centering
   \includegraphics[width=1.0\linewidth]{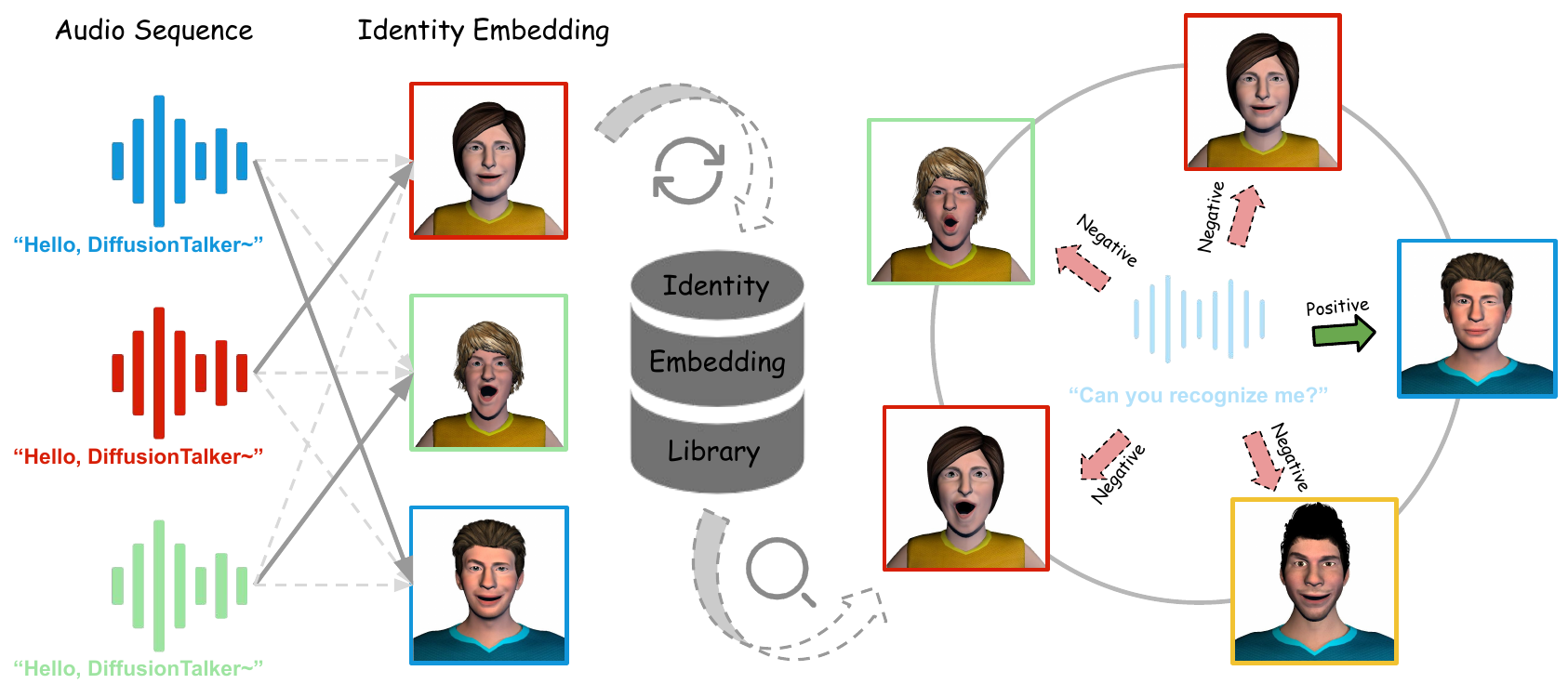}
   \caption{\textbf{Personalization Adapter.} During the training process, the personalization adapter updates the identity embedding library, whereas the inference process involves searching for the identity embedding within this library.}
   % \Description{}
   \label{fig:adaper}
   \vspace{-5mm}
\end{figure}

To bridge the gap between audio and talking identity matching, we propose the personalization adapter by contrastive learning, ultimately achieving the personalization of identity based on speech during inference. In the personalization adapter, we have constructed an identity embedding library based on the speakers, with each speaker corresponding to an identity embedding. We have configured these embeddings as trainable parameters during the training process. Both the audio sequence and its corresponding identity embedding are fed into the network as a training pair. The former is input into a pre-trained HuBERT model~\cite{hsu2021hubert} serving as the audio encoder to extract audio features, while the latter is received by an identity encoder with an MLP network to generate identity features. These identity embeddings and audio features are further passed on to the next layer of the network. Identity features and audio features are used to train the network with contrastive learning.

Regarding contrastive learning, we draw inspiration from ~\cite{radford2021learning}. However, in our approach, we have fixed the parameters of the audio encoder and only trained the identity encoder and learnable identity embeddings. The objective is to predict the correct relationship between identity embeddings and audio sequences. In terms of the specific details of contrastive learning, as shown in Fig.~\ref{fig:adaper}, we first apply L2 normalization to both the identity embeddings and audio features to normalize vector lengths, preserving only directional features. Secondly, we perform matrix multiplication on the normalized representations of both, and the resulting output is used to compute the cross-entropy loss with the labels. This loss is then used to update the parameters of the personalization adapter.

During inference, a given audio sequence is input into the audio encoder, generating an audio feature. This feature is then matrix-multiplied with the features of all embeddings in the identity embedding library. The embedding with the highest similarity is identified as the talking identity matching the input audio sequence.

\subsection{Knowledge Distillation}
To accelerate the inference of DiffusionTalker, we utilize knowledge distillation for fewer denoise steps. Inspired by progressive distillation~\cite{salimans2022progressive}, we introduce a student model $\hat{s}_{\eta}(\bm{w}_t)$ with $N$ steps to match the pre-trained teacher model $\hat{t}_{\theta}(\bm{w}_t)$ with $2N$ steps. $\bm{w}_{t}$ is defined as follows:

\begin{equation}
    \bm{w}_t = (\bm{a}, \bm{i}, \bm{x}_t, t)
\end{equation}
where $\bm{a}$ is the audio sequence, $\bm{i}$ is the identity embedding corresponding to $\bm{a}$, and $t$ is time step.
We first initialize the student model with a copy of the teacher model. Then we sample data and add noise to it according to the student DDPM $N$ steps. Before applying the student model $\hat{s}_{\eta}(\bm{w}_t)$ to denoise, we utilize the teacher model $\hat{t}_{\theta}(\bm{w}_t)$ to calculate the denoise output of $2N$ and $2N-1$ steps. To obtain a progressive distillation, we make one student DDPM step $N$ to match two teacher DDPM steps $2N$ and $2N-1$. 

Assume that the signal-to-noise ratio at time steps $\tau$ is $\alpha^2_\tau/\sigma^2_\tau$, then the corresponding data $\bm{x}_0$ added with noise $\bm{\epsilon}$ is $\bm{x}_\tau=\alpha_\tau \bm{x}_0 + \sigma_\tau \bm{\epsilon}, \bm{\epsilon} \sim N(0, I)$. For the teacher model, we can obtain noisy data $\bm{x}_{\tau'}$ at time step $\tau'=2\tau$ as the following: 
% \begin{equation}
%     \bm{w}_{\tau} = Concat(\bm{i}_e, \bm{a}_f, \bm{x}_{\tau}, \bm{\tau}_e)
% \end{equation}

\begin{equation}
    \bm{x}_{\tau'} = \alpha_{\tau'}\hat{t}_{\theta}(\bm{w}_\tau) + \frac{\sigma_{\tau'}}{\sigma_{\tau}}(\bm{x}_\tau-\alpha_\tau\hat{t}_{\theta}(\bm{w}_\tau))
\end{equation}
For $\tau''=2\tau-1$, noisy data $\bm{x}_{\tau''}$ can also be calculated with the above equation. As illustration in ~\cite{salimans2022progressive}, the final target $\widetilde{\bm{x}}$ from the teacher model can be formulated as:
\begin{equation}
    \widetilde{\bm{x}}=\frac{\bm{x}_{\tau''}-(\sigma_{\tau''}/\sigma_{\tau})\bm{x}_\tau}{\alpha_{\tau''}-(\sigma_{\tau''}/\sigma_{\tau})\alpha_\tau}
\end{equation}
We apply $\widetilde{\bm{x}}$ to supervise the student model along with other losses until convergence. Then we treat the student model with $N$ steps as the new teacher model and repeat the same progressive knowledge distillation process. Finally, DiffusionTalker can be distilled into several steps while maintaining high-generation quality.

\subsection{Training and Inference}
\subsubsection{Training}
In our training process, we randomly select a time step \( t \) from a uniform distribution ranging from 1 to \( T \). We then add noise to \( \bm{x}_0 \) for \( t \) steps to obtain \( \bm{x}_t \). Audio-identity training pairs are fed into the audio encoder and the identity encoder respectively to extract features. We conduct contrastive learning between the identity feature and the audio feature. The step \( t \) is input into the step encoder to obtain a step embedding, which is then aligned with other features. Subsequently, the identity embedding, audio feature, step embedding, and \( \bm{x}_t \) are concatenated and input into the talker decoder for denoising, producing \( \hat{\bm{x}} \). The model parameters are optimized based on the L2 loss between \( \hat{\bm{x}} \) and \( \bm{x}_0 \). The trainable components of this process include the identity embeddings, identity encoder, step encoder, and talker decoder, while the parameters of the audio encoder remain fixed. We propose five types of loss functions, namely, expression loss, lip loss, contrast loss, velocity loss, and distillation loss, each specifically targeted toward different constraint objectives. 

During the training process of the teacher model, the overall loss is formulated as:

\begin{equation}
    L_{tea} = \lambda_{1}L_{exp}+\lambda_{2}L_{lip}+\lambda_{3}L_{con}+\lambda_{4}L_{vel}
\end{equation}

During the knowledge distillation process of the student model, the overall loss is formulated as:  

\begin{equation}
    L_{stu} = L_{tea} +\lambda_{5}L_{dis}
\end{equation}
where $\lambda_{1}$ = 1.0, $\lambda_{2}$ = 1.0, $\lambda_{3}$ = 0.007, $\lambda_{4}$ = 0.5, $\lambda_{5}$ = 0.1 are fixed in all our experiments. The formulation of these loss functions will be explained below:

\noindent\textbf{Expression Loss.} 
Expression loss is a constraint applied to all facial parameters, and its definition is as follows:
\begin{equation}
    L_{exp} = \|\bm{\hat{x}}- \bm{x}_0\|^2 
\end{equation}
where $\bm{\hat{x}}$ is the prediction of facial parameters, $\bm{x}_0$ is the ground truth of animation. 

\noindent\textbf{Lip Loss.} 
In order to improve the accuracy of lip movement, we selected parameters related to the lips and applied lip loss to constrain these parameters.
\begin{equation}
    L_{lip} = \|\bm{\hat{x}}' - \bm{x}'_0\|^2 
\end{equation}
where $\bm{\hat{x}}'$ is the predicted parameters of lip area, $\bm{x}'_0$ is the label of lip animation. 

\noindent\textbf{Contrast Loss.} 
Contrast loss, as a target function for contrastive learning, is utilized to achieve the personalization of the talking identity. Its definition is as follows:
\begin{equation}
     L_{con}= -\frac{1}{M}\sum_{i=1}^M\log\frac{\sum_{t=0}^me^{h_i^\top  h_t/\tau}}{\sum_{j=0}^Me^{h_i^\top h_j/\tau}}, 
\end{equation}
where $h$ denotes the samples involved in contrastive learning, $M$ is the number of samples, $m$ is the number of positive samples corresponding to $i^{th}$ sample $h_i$, and $\tau$ is a temperature hyper-parameter.

\noindent\textbf{Velocity Loss.} 
We add the velocity loss from EmoTalk to enhance the smoothness between predicted animation frames. It is defined as follows:
\begin{equation}
    L_{vel} = \sum_{i=2}^{n} \|(\bm{\hat{f}}^i- \bm{\hat{f}}^{i-1})-(\bm{f}_{0}^i- \bm{f}_{0}^{i-1})\|^2 
\end{equation}
where $n$ is the number of total animation frames, and $\bm{f}$ is the animation parameters for a single frame.

\noindent\textbf{Distill Loss.}
We employ distill loss as the objective function for knowledge distillation, and its definition is as follows:
\begin{equation}
    L_{dis} = max(\frac{\alpha^{2}_{\tau}}{\sigma^{2}_{\tau}}, 1)\|\widetilde{\bm{x}} - \hat{s}_{\eta}(\bm{w}_\tau)\|^2 
\end{equation}

\subsubsection{Inference}
During the inference process, an input audio sequence will find its matched identity embedding in the personalization adapter, which serves as the talking identity. This embedding, along with the audio feature and other intermediate features, is concatenated and input into the talker decoder for denoising. After $N$ steps of denoising, the final facial parameters are predicted. This process, due to the inclusion of the talking identity, achieves user personalization.

% \subsection{Inference Procudure}

\section{Experiments}

\subsection{Experimental Settings}
\noindent\textbf{Datasets.} 
% \subsubsection{Datasets}
In the field of speech-driven 3D facial animation, the two most common approaches are blendshape-based facial animation~\cite{lewis2014practice} and vertex-based facial animation. Correspondingly, the datasets used are also categorized into blendshape-based and vertex-based datasets. Motivated by the objective of speed enhancement in DiffusionTalker, we trained our DiffusionTalker on the blendshape-based BEAT dataset~\cite{Liu2022beat} for its fewer parameters. BEAT are collected by Apple ARKit with 52 blendshape coefficients. It consists of voice recordings and corresponding blendshape-based facial animations from 30 different individuals. Each person says the same content but with their unique speaking style.We utilize a subset with about 32 hours of audio data from the full dataset. We further segmented the audio data into 11,427 sequences, each with a duration of 10 seconds. We follow the data preprocessing in FaceDiffuser~\cite{stan2023facediffuser}.

We also employed VOCASET~\cite{cudeiro2019capture} as a zero-shot test set to more effectively evaluate the robustness of DiffusionTalker. VOCASET, a vertex-based dataset, comprises 480 facial animation sequences and the corresponding speech data, featuring 12 subjects. Each facial animation sequence consists of 5,023 vertices, multiplied by 3 to represent spatial coordinates in the $x$, $y$, and $z$ dimensions. The duration of each speech data segment is approximately 3-4 seconds. For testing, we adopted the Blendshape conversion method from EmoTalk~\cite{peng2023emotalk}, enabling the transformation of VOCASET into 52 blendshape coefficients.

\noindent\textbf{Implementation Details.} 
% \subsubsection{Implementation Details}
We follow the implementation details specified in FaceDiffuser~\cite{stan2023facediffuser}.
For model architectures, we utilize the pretrained HuBERT~\cite{hsu2021hubert} as an audio encoder to extract audio features. The dimension of identity embeddings is set to 32 and is sent to the identity encoder. We further concatenate the encoded features and employ a two-layer GRU. The dimension of the hidden layer is set as 256. 
To train the whole pipeline, we utilize the Adam optimizer~\cite{kingma2014adam} with $(\beta_1, \beta_2) = (0.9, 0.999)$ to update parameters, and the learning rate is set to $1e^{-4}$. We set a linear $\beta$ schedule for 32 steps for teacher models. We first train the teacher model with 50 epochs. Then we copy the weights of the teacher model into the student model and apply the proposed knowledge distillation techniques to train the student model with half-time steps. We iteratively set the student model as the teacher model to further reduce sampling time steps. All experiments are conducted on NVIDIA V100 GPUs. More implementation details are reported in the supplementary material.

\noindent\textbf{Evaluation Metrics.} 
% \subsubsection{Performance Metrics}
To quantitatively assess the performance and robustness of DiffusionTalker, we employed four key evaluation metrics: MBE, LBE, FDD, and ITF.

(MBE) Mean Blendshape Error measures the average Euclidean distance between the predicted blendshape coefficients and the ground truth. Lower MBE values indicate higher accuracy, as they represent smaller deviations from the ground truth.

(LBE) Lip Blendshape Error measures the discrepancy between the predicted and ground truth blendshape coefficients associated with the lip region.

(FDD) Facial Dynamics Deviation~\cite{xing2023codetalker} measures the deviation in the dynamics of the upper face region between the generated sequence and the ground truth.

(ITF) Inference Time per Frame is defined as the inference time per audio frame, calculated by dividing the total inference time by the total number of audio frames. This metric is employed to quantitatively measure and compare the inference speed of models.

\subsection{Experimental Results}

\subsubsection{Quantitative Evaluation}
We trained six DiffusionTalker models from scratch on the BEAT dataset, with steps set covering 256, 128, 64, 32, 16, and 8 respectively. Subsequently, we employed the knowledge distillation approach where a 32-step model served as the teacher model to train a 16-step student model. This 16-step student model was then used as a new teacher to distill an 8-step student model.

In our quantitative experiments, we conduct three sets of tests. In the first set, we assess and compare the performance metrics of all nine models, ranging from 256 to 8 steps. In the second set, we retrain FaceDiffuser on the BEAT dataset. EmoTalk, FaceDiffuser, and the 8-step distilled DiffusionTalker are tested on BEAT, and metrics such as MBE, LBE, FDD, and ITF are reported. The reason for selecting EmoTalk and FaceDiffuser as baselines is that other methods do not release the training code on blendshape-based datasets. In the third set, we retrain FaceDiffuser on the VOCASET and conduct an evaluation. We use the test results from EmoTalk's paper, including methods like FaceFormer and EmoTalk on VOCASET.  Our 8-step distilled DiffusionTalker trained on BEAT is directly tested on VOCASET and compared with other methods. In the third set, We focus on lip accuracy and inference time, hence the LVE and ITF metrics are reported.

\begin{table}[t]
    \centering
    \caption{\textbf{The performance and inference speed of our model vary with different step counts.} Specifically, the inference speed of the 8-step model is enhanced when using NVIDIA TensorRT.}
    \resizebox{\linewidth}{!}{
    \begin{tabular}{c|cc|ccc}
    \toprule
        Dataset & Steps & Distilled & LBE$\downarrow$ & $\|$FDD $\|$$\downarrow$ &ITF($10^{-4}$s)$\downarrow$\\
    \midrule
         % & 500 & 4.704  \\
         \multirow{9}*{BEAT} & 256 & \ding{55} &0.1050 & 0.1693 & 255.5 \\
         \multirow{9}*{} & 128 & \ding{55} &0.1059 & 0.1445 & 132.4 \\
         \multirow{9}*{} & 64 & \ding{55} &0.1063 & 0.1517 & 65.3 \\
         \multirow{9}*{} & 32 & \ding{55} &0.1053 & 0.1647 & 32.2 \\
         \multirow{9}*{} & 16 & \ding{55} &0.1097 & 0.1656 &  16.8 \\
        \multirow{9}*{} & 8 & \ding{55} &0.1177 & 0.1709 & 7.0 \\
    \cline{2-6}
         \multirow{9}*{} & 16 & \checkmark &\textbf{0.0957} & \textbf{0.1433} & 16.8 \\
         \multirow{9}*{} & 8 & \checkmark &0.0969 & 0.1558 & 7.1 \\
         
         \multirow{9}*{} & 8(TRT) & \checkmark &0.0975 & 0.1563 & \textbf{3.9} \\
         % \midrule
         % & w/o Diffusion & \textbf{2.6963} & \textbf{1.5924} & 2.0553 \\
         % UUDaMM & B-FaceDiffuser & 3.4479 & 1.6671 & \textbf{1.7752}\\
    \bottomrule
    \end{tabular}
    } %resize end
    % \Description{}
    \label{tab:eval_on_various_steps}
\end{table}

\begin{table}[t]
    \centering
    \caption{\textbf{The quantitative results evaluated on the 3D blendshape dataset.} Various models were tested on the BEAT dataset, with the best results highlighted in bold. Among them, FaceDiffuser is a 1000-step model.  Our model demonstrates superior performance in MBE, LBE, and ITF, except for the FDD metric.}
    \resizebox{\linewidth}{!}{
    \begin{tabular}{c|c|cccc}
    \toprule
        Dataset & Method & MBE $\downarrow$ & LBE $\downarrow$ & $\|$FDD $\|$$\downarrow$  & ITF($10^{-4}$s)$\downarrow$\\
         % &  &x$10^{-3}$ & x$10^{-4}$  & x$10^{-5}$& x$10^{-3}$\\
         % &  & mm &  mm  & mm & mm\\
    \midrule
         % & FaceFormer & - & - & - & -\\
         \multirow{3}*{} & EmoTalk & 1.0537 & 0.2628 & 0.0751 & 9.3\\
         \multirow{3}*{BEAT} & FaceDiffuser & 0.4332 & 0.1071 & 0.1786 & 1000.9 \\
         \multirow{3}*{} & FaceDiffuser-8 & 0.8372 & 0.2115 & \textbf{0.0589} & 8.2 \\
         \multirow{3}*{} & Ours-8 & \textbf{0.4094} & \textbf{0.0969} & 0.1558 & \textbf{7.1} \\
         
    \bottomrule
    \end{tabular}
    } %resize end

    \label{tab:eval_on_beat}
\end{table}

\begin{table}[t]
    \centering
    \caption{\textbf{The quantitative results evaluated on the 3D vertex dataset.} Various models were tested on the VOCASET dataset, with the best results highlighted in bold. Our model, without being trained on the VOCASET dataset, demonstrated optimal performance in terms of the LVE and ITF metrics under the zero-shot condition.}
    \resizebox{\linewidth}{!}{
    \begin{tabular}{c|c|ccc}
    \toprule
        Dataset & Method & LVE $\downarrow$& ITF($10^{-4}$s)$\downarrow$& Zero-Shot\\
    \midrule
         % \multirow{5}*{VOCASET}  & VOCA & 4.704 & 4.2 & \ding{55} \\
         % & MeshTalk & 4.513 & - & \ding{55} \\
         \multirow{5}*{VOCASET}  & FaceFormer & 4.418 & 54.2 & \ding{55} \\
         \multirow{5}*{}  & EmoTalk & 4.134 & 9.3 & \checkmark \\
         \multirow{5}*{}  & FaceDiffuser-8 & 4.713 & 8.2 &  \ding{55} \\
         \multirow{5}*{}  & Ours-8 & \textbf{4.054} & \textbf{7.1} &  \checkmark \\
         % \midrule
         % & w/o Diffusion & \textbf{2.6963} & \textbf{1.5924} & 2.0553 \\
         % UUDaMM & B-FaceDiffuser & 3.4479 & 1.6671 & \textbf{1.7752}\\
    \bottomrule
    \end{tabular}
    } %resize end
    \label{tab:eval_on_vocaset}
\end{table}

% DiffusionTalker is primarily focused on enhancing inference speed. By employing additional constraints like lip loss and velocity loss, the model significantly reduces the denoising steps required in a typical Diffusion model from over a hundred to just 32, while maintaining its original performance. Subsequently, we applied knowledge distillation techniques to progressively refine the 32-step Diffusion model into an 8-step version, achieving a substantial increase in inference speed. 

\begin{figure}[t]
  \centering
   \includegraphics[width=0.9\linewidth]{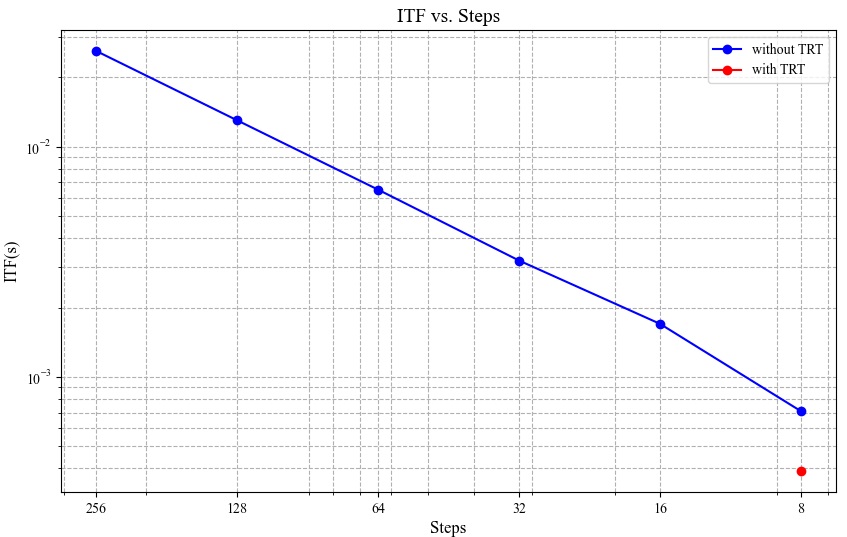}
   \caption{\textbf{ITF vs. Steps.} The inference speeds of the model with and without NVIDIA TensorRT vary under different step counts.}
   % \Description{}
   \label{fig:steps_comp}
   \vspace{-5mm}
\end{figure}

\noindent \textbf{Inference Speed.} As shown in Tab.~\ref{tab:eval_on_various_steps}, the ITF decreases as the number of steps reduces, indicating that a decrease in steps leads to an increase in the inference speed. This observation aligns with our general understanding. Consequently, we plotted ITF against the number of steps, as depicted in Fig.~\ref{fig:steps_comp}, which demonstrates a direct proportional relationship between ITF and steps, further indicating an inverse relationship between model inference speed and the number of steps. The last row of Tab.~\ref{tab:eval_on_various_steps} highlights that upon integrating the 8-step distilled model with NVIDIA TensorRT, the ITF decreased from $7.1 \times 10^{-4}s$ to $3.9 \times 10^{-4}s$, resulting in an 82.1\% increase in inference speed.

We compared the 8-step distilled model on the BEAT and VOCASET test sets with other methods, as illustrated in Tab.~\ref{tab:eval_on_beat} and~\ref{tab:eval_on_vocaset}. The model exhibited the best ITF metric in both cases, indicating the fastest inference speed. Our method achieves a similar speed compared with FaceDiffuser-8, however, our quantitative results are much better thanks to the personalization ability.

%This can be attributed to: (1) directly reducing the steps in the Diffusion model or reducing them via knowledge distillation significantly enhances inference speed; (2) optimizing model parameters using NVIDIA TensorRT further accelerates model inference.
% As indicated in Tab.~\ref{tab:eval_on_various_steps}, the 8-step model achieves an inference speed of $7.1 \times 10^{-4}$ seconds per frame. Building upon the 8-step model, we utilized NVIDIA TensorRT, a ready-to-use inference acceleration toolkit, to optimize the neural network, thereby reducing inference time. Notably, the inference speed of the 8-step model with TensorRT shows an improvement of 82.1\% compared to the 8-step model without TensorRT.
% As illustrated in Fig.~\ref{fig:steps_comp}, within the range of 256 to 8 steps, there is a direct proportional relationship between inference time and the number of steps: the fewer the steps, the less time required for inference, resulting in faster processing speeds. 

% As demonstrated in Tables \ref{tab:eval_on_beat} and \ref{tab:eval_on_vocaset}, our 8-step model (Ours-8) achieves an inference speed of $3.9 \times 10^{-4}$ seconds per frame, making it the fastest among all compared methods and approximately twice as fast as EmoTalk. Consequently, in terms of inference speed, DiffusionTalker has established itself as a state-of-the-art solution.

\noindent \textbf{Fidelity and Accuracy.} Lip movement precision is a crucial factor for evaluating fidelity and accuracy in the speech-driven 3D facial animation field. As indicated in Tab.~\ref{tab:eval_on_various_steps}, without knowledge distillation, the LBE and FDD metrics slightly increase as the number of steps decreases. This suggests that reducing the steps in a diffusion model can lead to a decrease in accuracy. During the transition from 16 to 8 steps with knowledge distillation, this trend persists. However, models obtained through knowledge distillation at identical step counts (16 or 8) significantly outperform their non-distilled counterparts in terms of LBE and FDD. This implies that models derived via knowledge distillation exhibit enhanced accuracy. This trend can be attributed to (1) a reduction in the model's steps, leading to fewer sampling steps and potentially inadequate de-noising, affecting precision; (2) during knowledge distillation, the student model learns soft labels from the teacher model, encompassing the latter's understanding of the data, which might aid the student model in better generalizing to unseen data; (3) using the soft outputs of the teacher model as targets may provide a regularizing effect, helping to prevent overfitting in the student model. 

Tab.~\ref{tab:eval_on_beat} and~\ref{tab:eval_on_vocaset} demonstrate that our approach achieves state-of-the-art performance in MBE and LBE metrics, both on BEAT and VOCASET. However, our model exhibits suboptimal performance in the FDD metric. This suggests a weaker inference capability in terms of facial dynamics deviation, particularly in the upper face region. This limitation could stem from the model's excessive focus on lip accuracy constraints, leading to a comparatively lesser emphasis on dynamics of areas with weak correlation to speech. 
%This success can be attributed to: (1) our use of target functions like lip loss and velocity loss, which may impose stricter constraints on facial and lip accuracy; (2) our model, obtained through knowledge distillation, likely exhibits superior generalization capabilities compared to models trained with hard labels; (3) the inclusion of identity embeddings corresponding to speech as a conditioning factor for de-noising in our model, which may enable more precise control of the generated blendshape information. 

\noindent \textbf{Identity Matching.} Identity matching is a crucial component for achieving personalization ability. We tested our 8-step distilled model on the whole BEAT dataset. As shown in Tab.~\ref{tab:identity_match_acc}, the results indicate that our contrastive learning approach effectively integrates the audio and identity modalities, achieving high accuracy in identity matching. 

\begin{table}[t]
    \centering
    \caption{\textbf{Identity matching results.}}
    \resizebox{0.8\linewidth}{!}{
    \begin{tabular}{c|ccc}
    \toprule
        Metrics & Precision & Recall & F1 Score\\
    \midrule
        Results(\%) & 99.996 & 99.986 & 99.991 \\
    \bottomrule
    \end{tabular}
    }
    \label{tab:identity_match_acc}
\end{table}

% \begin{table}
%     \centering
%     \caption{
%         \label{tab:identity_match_acc}  
%         Identity matching results.
%     }
%     \begin{tblr}{
%         colspec={l | c c c }
%     }
%     \hline
%     \SetCell[r=1,c=1]{c} Metrics & Precision & Recall & F1 Score  \\
%     \hline
%     \makebox[\nameblob][l]{Results(\%)}  & 99.996 & 99.986 & 99.991 \\
%     \hline
%     \end{tblr}    
% \end{table}
% \paragraph{\textbf{Facial Expression Diversity}}

\begin{table}[t]
    \centering
    \caption{\textbf{Ablation study for our 8-step model.} We conducted an evaluation of key components within our model by assessing the impact on MBE and LBE metrics in their absence.}
    \resizebox{0.8\linewidth}{!}{
    \begin{tabular}{c|ccc}
    \toprule
        Ablation Settings & MBE$\downarrow$ & LBE $\downarrow$& \\
    \midrule
          Ours & \textbf{0.4094} & \textbf{0.0969}  \\
    \midrule
          w/o knowledge distillation & 0.4545 & 0.1177  \\
          w/o $L_{lip}$ loss & 0.4222 & 0.1037  \\
          w/o $L_{vel}$ loss & 0.4414 & 0.1129  \\
          w/o identity embedding & 0.4306 & 0.1034   \\
          % $L_{vel}$ loss with 2nd type & 0.4874 & 0.1185  \\
    \bottomrule
    \end{tabular}
    } %resize end
    \label{tab:ablation_sduty}
\end{table}

\noindent \textbf{Effectiveness of Each Component.} We conducted a study to evaluate the impact of each component. As presented in Tab.~\ref{tab:ablation_sduty}, both metrics increase without knowledge distillation. The $L_{lip}$ loss specifically reduces the error rate in the lip region, while the $L_{vel}$ loss focuses on the differences in overall blendshapes between frames. The absence of identity embedding could lead to a decrease in performance. This study shows the effectiveness of each component.

\subsubsection{Qualitative Evaluation}

% \begin{figure*}[h]
% \vspace{-10pt}
%     \centering
%     \includegraphics[width=0.8\textwidth]{figs/visualization_v2.pdf}
%     \vspace{-5pt}
%     \caption{\textbf{Comparisons with other works on various talking identities.}
%     Comparisons with other works on various talking identities.}
%     \label{fig:method}
%     \vspace{-10pt}
% \end{figure*}

\begin{figure}[t]
  \centering
   \includegraphics[width=1.0\linewidth]{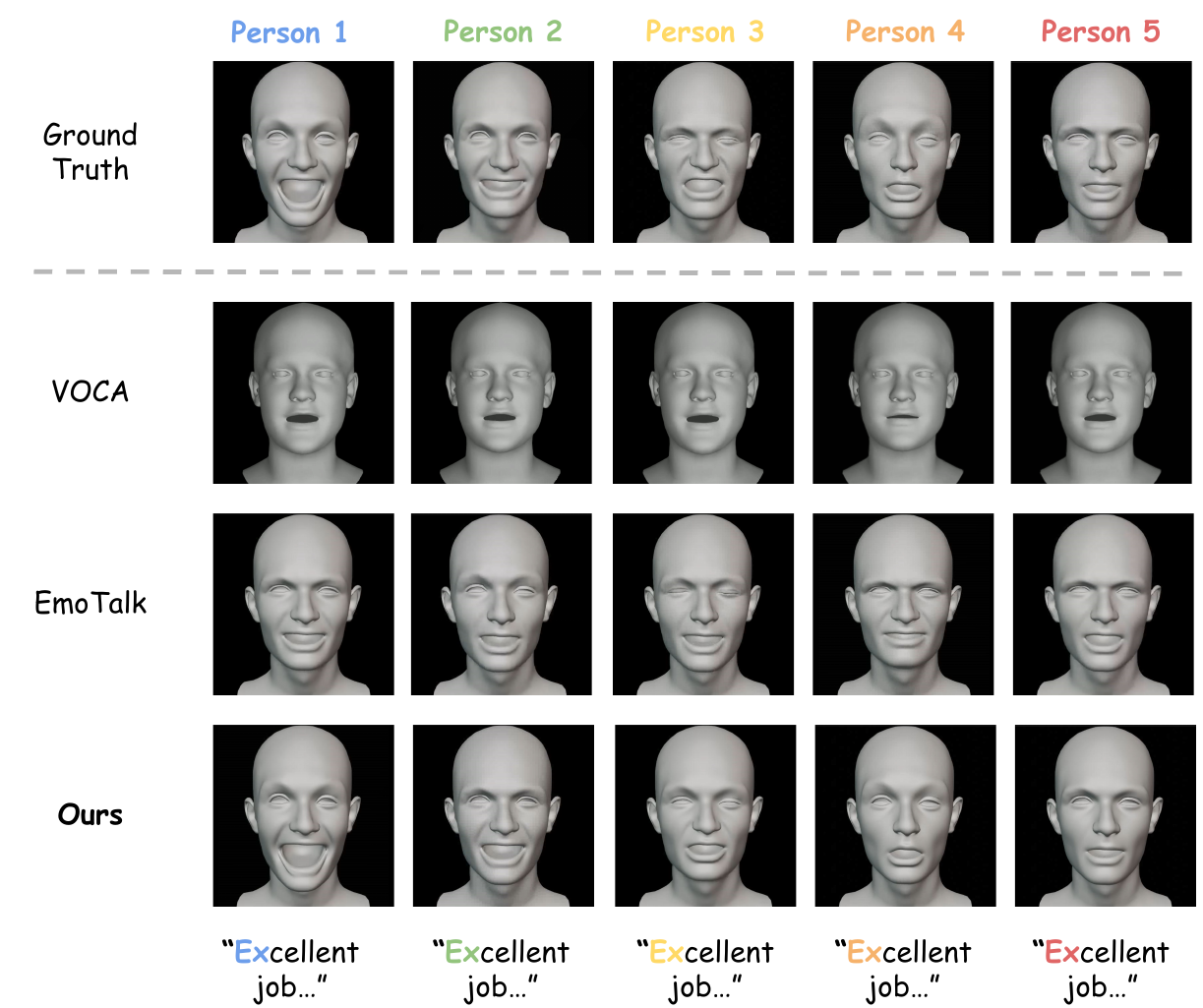}
   \caption{\textbf{Qualitative comparisons on various talking identities with other methods.} We selected audio sequences from five different individuals as tests. In each of the cases representing different talking identities, we emitted the $/'\varepsilon ks/$
 sound in the word "excellent." We conducted qualitative comparisons with ground truth respectively using VOCA, EmoTalk, and DiffusionTalker.}
   % \Description{}
   \label{fig:visualization}
   \vspace{-5mm}
\end{figure}

\noindent \textbf{Talking Identity Personalization.} As shown in Fig.~\ref{fig:visualization}, we choose five audio sequences from different people. For instance, Person 1 consistently speaks with exaggerated facial expressions, displaying larger facial movement amplitudes. Compared with other methods, DiffusionTalker generates facial animations that are closer to the ground truth, effectively demonstrating the personal talking identity. On the other hand, VOCA and EmoTalk are less proficient in this regard, failing to capture and display the unique speaking expression habits and styles of individuals.The qualitative comparisons tested on VOCASET and user study are reported in the supplementary materials.

% \paragraph{\textbf{Fidelity and Accuracy}}
\begin{figure}[t]
  \centering
   \includegraphics[width=1.0\linewidth]{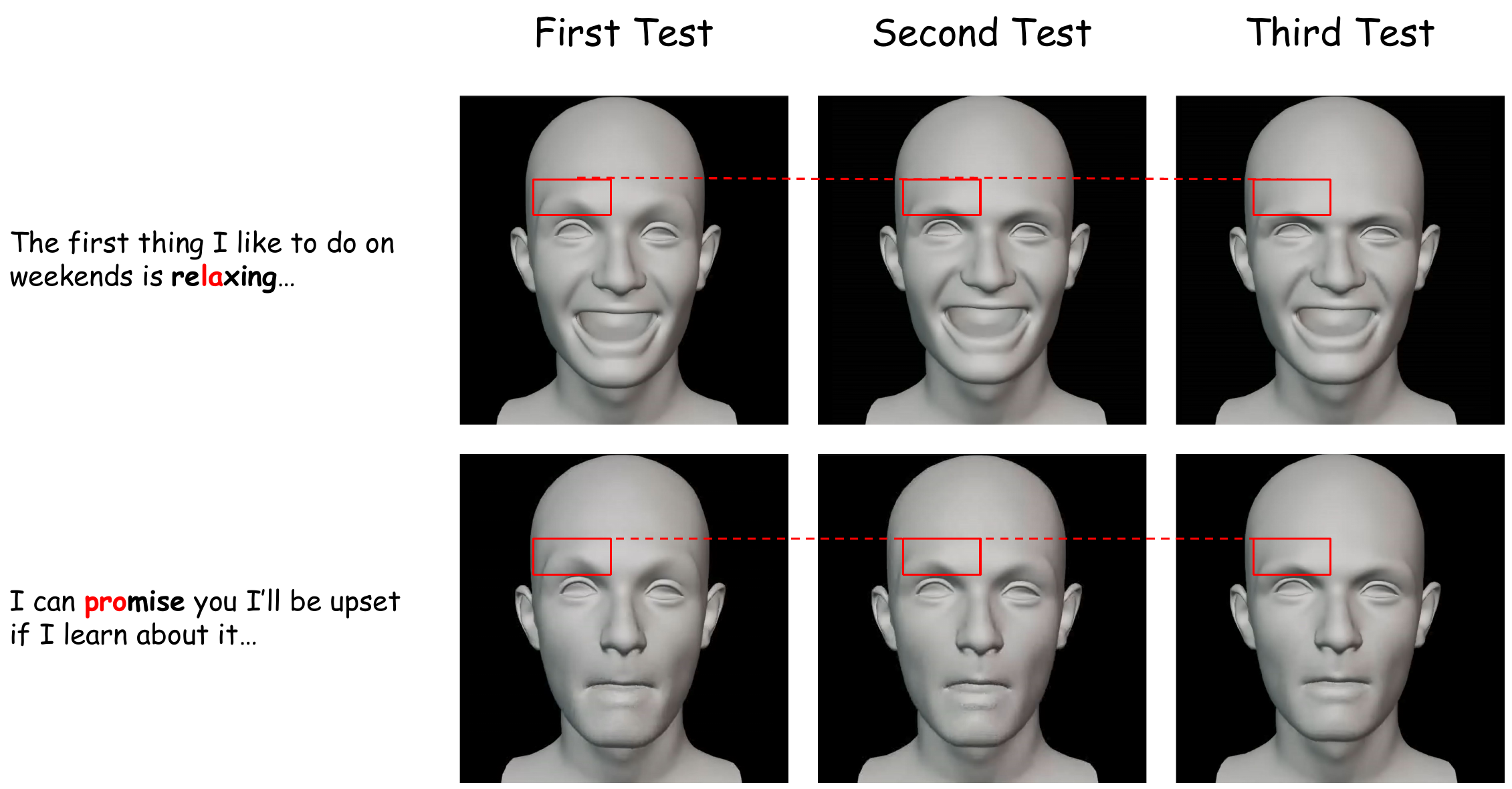}
   \caption{\textbf{Facial Expression Diversity.} The location of eyebrowns is non-deterministic among various tests. }
   % \Description{}
   \label{fig:diversity}
   \vspace{-2mm}
\end{figure}

\noindent \textbf{Facial Expression Diversity.} As illustrated in Fig.~\ref{fig:diversity}. Under the same input audio and identity, we conducted three inferences. It can be observed that, in the same audio frame, the corresponding animation frames show noticeable differences in the position of the eyebrows, demonstrating the non-deterministic fact of speech-driven 3D face animation.

% Maybe in supplement material
% \paragraph{\textbf{User study}}

% \paragraph{\textbf{Knowledge Distillation}}

% \paragraph{\textbf{Audio Encoder}}

% \paragraph{\textbf{Facial Decoder}}

% \paragraph{\textbf{Loss coefficient}}

% \paragraph{\textbf{Identity Embedding Dimension}}

% \paragraph{\textbf{Identity Encoder}}

% \subsection{Discussion} \cref{tab:obj_vertex_data} and \cref{tab:obj_blendshape_data} show the objective results for V-FaceDiffuser and B-FaceDiffuser respectively. Our approach performs better than all the other methods on all the objective metrics for the BIWI dataset. For the Multiface dataset, ours perform the best on all objective metrics but FDD, for which FaceFormer performs slightly better. For the blendshape based datasets, we cannot compare our method with state-of-the-art methods as mentioned earlier. Instead, we compare the diffusion  MBE and LBE are higher than the baseline model as our approach encourages randomness and non-determinism while more resembling the upper face variation observed in the ground truth with lower FDD value. Furthermore, the diversity for B-FaceDiffuser is evaluated qualitatively in \cref{sec:qualitative}. 

\section{Conclusion}
To generate personalized speech-driven 3D facial animation in a short time, we propose a diffusion-based method named DiffusionTalker. The paper applies contrastive learning between encoded audio features and learnable talking identity to aggregate personal information. To further accelerate the inference time, we distill dense knowledge from a teacher model with huge sampling steps into a student model with much fewer sampling steps while maintaining competitive accuracy. Finally, we conduct detailed experiments on two datasets BEAT and VOCASET, demonstrating the effectiveness of our proposed method.In the future, we will explore more natural 3D facial animation, and personalized facial texture features based on speech show promising potential.

%Although the diffusion model is effective enough for speech-driven facial animation, it still requires large datasets for better performance and convergence. It is difficult for us to collect plenty of real data for diffusion model training. Constrained by the identity numbers of datasets, DiffusionTalker may fail to cover whose voice is far from learned identities in the database. Despite limitations, DiffusionTalker has demonstrated that speech-driven 3D facial animation can benefit from the characteristics of the diffusion model. In the future, we will further explore the nature of the diffusion model for aiding 3D animations. 
{
    \small
    \bibliographystyle{ieeenat_fullname}
    \bibliography{main}
}


\end{document}